# Main manuscript for

# Robot motor learning shows emergence of frequency-modulated, robust swimming with an invariant Strouhal number

Hankun Deng[1], Donghao Li[1], Colin Nitroy[1], Andrew Wertz[1], Shashank Priya[2], Bo Cheng[1]

[1]Department of Mechanical Engineering, The Pennsylvania State University, University Park, PA 16802; [2]Department of Material Science and Engineering, The Pennsylvania State University, University Park, PA 16802.

*Corresponding author: Bo Cheng

Email: buc10@psu.edu

**This PDF file includes:**

    Main Text
    Figures 1 to 5

**Abstract**

Fish locomotion emerges from a diversity of interactions among deformable structures, surrounding fluids and neuromuscular activations, i.e., fluid-structure interactions (FSI) controlled by fish's motor systems. Previous studies suggested that such motor-controlled FSI may possess embodied traits. However, their implications in motor learning, neuromuscular control, gait generation, and swimming performance remain to be uncovered. Using robot models, we studied how swimming behaviours emerged from the FSI and the embodied traits. We developed modular robots with various designs and used Central Pattern Generators (CPGs) to control the torque acting on robot body. We used reinforcement learning to learn CPG parameters to maximize the swimming speed. The results showed that motor frequency converged faster than other parameters, and the emergent swimming gaits were robust against disruptions applied to motor control. For all robots and frequencies tested, swimming speed was proportional to the mean undulation velocity of body and caudal-fin combined, yielding an invariant, undulation-based Strouhal number. The Strouhal number also revealed two fundamental classes of undulatory swimming in both biological and robotic fishes. The robot actuators also demonstrated diverse functions as motors, virtual springs, and virtual masses. These results provide novel insights into the embodied traits of motor-controlled FSI for fish-inspired locomotion.

**Main text**

## 1. Introduction

The diversification of fish locomotion rests on 530 million years of evolution that successfully explored various forms of fluid-structure interaction (FSI) for underwater locomotion [1,2]. The most common forms of fish locomotion, i.e., body and/or caudal-fin (BCF) swimming [3], include active structures such as fins and elongated bodies; they are deformed by an array of distributed muscle forces, leading to interactions with the surrounding fluids and their passive deformation by hydrodynamic forces, forming a FSI loop (figure 1*ai*) [4]. Undulatory swimming gaits emerge from such FSI under rhythmic control of fishes' motor system, which includes central and peripheral neural systems that combine feedforward control patterns with peripheral sensory feedback [5]. This combination offers high robustness of rhythmogenesis and gait coordination against neural disruptions, as evidenced by a recent study using a robot to emulate lamprey neuromechanics [6].

The interplay between fish's motor system and physics of FSI is remarkable complex, especially considering the diversity in fishes' morphological traits and swimming behaviours [7]. However, previous studies have hinted that such motor-controlled FSI of fish swimming may possess embodied traits that are independent of body size, morphologies and neuromuscular control. For example, most BCF fish species exhibit similar midline kinematics during steady swimming and share major kinematic features, suggesting the existence of unifying locomotor hydrodynamic mechanisms across a variety of body morphologies [8]. Studies have shown that swimming speed of multiple fish species (e.g., dace [9], green jack [10], and odontocete cetaceans [11]) is proportional to caudal-fin beating frequency, suggesting simplicity in speed control. The Strouhal number, which characterizes the wake patterns and hydrodynamics of undulatory swimmers [12], is typically in a narrow band for a single species swimming at various speed [13], presumably due to the relatively small variations in caudal-fin beating amplitude and the proportionality between speed and caudal-fin frequency [9]. Furthermore, fishes actively modulate their body stiffness at various undulation frequencies (thereby changing the properties of FSI), which enables them to match their body's natural frequency with the undulation frequency of the caudal-fin and therefore minimizes the mechanical cost of bending [14]. All these findings suggest that there exist inherent properties embodied in the FSI that govern fish or fish-inspired locomotion, and they may offer simplicity and high predictability for gait generation and swimming performance.

However, the FSI-embodied traits in fish or fish-inspired robotic swimming have not yet been subjected to rigorous, mechanistic studies, and current understandings linger on the scattered observations that alluded to these traits [15]. In particular, the implications of FSI-embodied traits in motor learning, neuromuscular control, gait generation, and swimming performance have not been systematically tested [16] and many questions remain to be addressed. For example, is the observed proportionality between speed and caudal-fin frequency a result of robustness in gait generation and swimming speed? Does it translate to simplicity in motor learning? i.e., fast convergence of frequency? Can we also establish a trackable, quantitative mapping between motor control, swimming gaits, and speed? Additionally, whether and how do the muscle functions vary with motor control frequency or speed? Answering the above questions is challenging if not infeasible based on direct measurements in living fishes, partly due to the challenges in training the fishes to swimming at various speed and in vivo measurement of muscle activation [17]. However, understanding the embodied traits has critical importance for advancing the biomechanics of fish swimming and their informed, principled mimicry in bio-inspired robotics.

Robot models (sometimes referred to as robophysical models) offer a unique opportunity to provide quantitative results that address the above questions [18]. Using robophysical models, not only can we generate variations in morphology, but we also have full motor control and the ability to monitor internal variables and swimming gaits. This allows for systematic testing and validation of embodied properties and their implications. Moreover, the discoveries made in robotics can inform the study of living organisms, leading to new hypotheses and insights into their functions (i.e., robotics inspired biology [19]).

In this work, we used robophysical models to systematically study the embodied properties and their implications in motor learning, neuromuscular control, gait emergence, and swimming speed (figure 1*a*). We designed and assembled modular robots for BCF swimming, named µBot (see materials and methods, figure 1*b*). The robots were directly torque-controlled by electromagnetic motors receiving voltage inputs generated by Central Pattern Generators (CPGs) (i.e., the motor programs; see materials and methods, figure 1*c*). We generated variations in the robot designs. For each robot, we first used Reinforcement Learning (RL) to optimize the CPG parameters for maximizing the swimming performance (forward or backward swimming speed, figure 1*aiii*). Note that, rather than being optimized directly, the swimming gaits emerged from the combined motor control and FSI (figure 1*ai* and 1*aii*). Next, we measured the frequency response of the robots based on the optimized motor programs, and analysed the changes in swimming gaits, speed, actuator work loops and cost of transport. Finally, we applied disruptions to the robot motor control and evaluated the robustness in the gait generation by performing sensitivity analysis on the emergent swimming gaits and speed.

## 2. Materials and methods

### 2.1. µBot design and experimental setup

µBots were developed as modular, robophysical models for BCF swimming. They were easily modifiable in morphologies, simple in assembling, compact in size, and durable for experimental testing [20,21]. Therefore, they enabled rapid prototyping with different design configurations (such as body length, shape, and stiffness) and allowed for extensive experimental learning trials to be conducted in controlled lab settings. µBot modules included a head module, body segment modules, a caudal-fin module, and rubber suit modules. Both head and body modules included electromagnetic actuators. In this work, we assembled and tested µBots with various body actuators or Degrees-of-Freedoms (µBot-DoFs, DoFs = 2, 4, 6) as the number of body segments is one of the most important parameters that determine the body kinematics [22]. A fully assembled µBot had 2 cm in width and 3 cm in depth, while its length and weight depended on the number of body segments or DoFs (µBot-2: 117 mm, 39 g; µBot-4: 170 mm, 59 g; µBot-6: 224 mm, 78 g). A µBot-4 with the caudal-fin removed was also experimentally studied. More details of the robot design and assembly can be found in the electronic supplementary material.

The experimental platform for µBots also included a water tank (58 cm $W$ × 56 cm $H$ × 305 cm $L$), a monochrome camera (acA2000-165umNIR, Basler AG Inc, Ahrensburg, Germany) with a 760 nm filter, IR light sources, and a motor controller (figure 1*aiii*). The laptop was used to send motor signals to the robot through the motor controller and also used for robot motion capturing. The serial communication for robot control and motion capturing was achieved with MATLAB 2020a (MathWorks, Natick, MA, USA) and was running at 50 Hz. Details of image processing and extraction of swimming gaits are provided in the electronic supplementary material.

**2.2. μBot motor control and experimental learning**

The robot motor control signals were generated by a Central Pattern Generator (CPG) which is frequently used in swimming robots [23,24]. The CPG we used was a dual-neuron model proposed by Matsuoka [25]. The CPG network for μBot-2 is illustrated in figure 1c. Each actuator was controlled by a CPG module, including two neurons inhibiting each other, while more modules were added for μBot-4 and μBot-6. The equations of each neuron were represented as,

$$\begin{aligned}
&\tau \dot{U}_{i,j} + U_{i,j} = E_i - \beta_i V_{i,j} - \alpha y_{i,3-j} + \omega y_{i-1,j} \\
&\tau \dot{V}_{i,j} + V_{i,j} = y_{i,j} \\
&y_{i,j} = max(0, U_{i,j}) \\
&y_{i,out} = y_{i,1} - y_{i,2} \\
&i = 1,2,\ldots,n;\ j = 1,2
\end{aligned} \quad (2.1)$$

where $U$ and $V$ are the states; $n$ is CPG module number; $\tau$ is the time constant; $E_i$ represents external stimulus for $i^{th}$ module; $\beta_i$ is adaption coefficient of $i^{th}$ module; $\alpha$ is mutual inhibition weight; $\omega$ is the inter-module connection weight of the neuron; $y_{i,out}$ is the output of the $i^{th}$ CPG module [20].

The rhythmic motor signals generated by the CPG can be characterized by their frequency, intensities, and intersegmental phases. In the CPG, $\tau$, $\beta_i$, and $E_i$ determine frequency, intersegmental phases, and intensities of motor control, respectively. To simplify the learning process, all the other parameters in the CPG were fixed (table S1). The parameter vector to be learned was $\boldsymbol{v} = [\tau, \beta_2, \ldots, \beta_n, E_1, \ldots E_n]$.

In our previous work, we used parameter exploring policy gradient (PGPE) reinforcement learning method for experimental motor learning [20]. Although PGPE showed good convergence, it required empirical tuning of the heuristic learning rate to reduce the number of episodes. To reduce the number of empirically-tuned hyper-parameters, here we used EM-based Policy Hyper-Parameter Exploration (EPHE) algorithm which integrates PGPE with EM-based updates that maximize a lower bound of the expected return in each parameter updating [26]. In EPHE, the policy parameter $\boldsymbol{v}$ is sampled from normal distribution $N(\boldsymbol{v} \mid \boldsymbol{h})$, where $\boldsymbol{h}$ is the hyperparameter vector governing the distribution of policy parameters. The hyperparameter vector $\boldsymbol{h} = [\boldsymbol{\mu}\ \boldsymbol{\sigma}]^T$ includes **μ** and **σ** for the mean and standard deviation of normal distribution of CPG parameters vector **v**. At each episode, $M$ policy parameters $v^m$ were sampled from $N(v \mid h)$. Then $K$ best parameters were selected from the sorted reward $R(v^m)$. The hyperparameter vector **h** was updated as,

$$\mu = \frac{\sum_{k=1}^{K} [R(v^k) v^k]}{\sum_{k=1}^{K} R(v^k)}, \tag{2.2}$$

$$\sigma = \sqrt{\frac{\sum_{k=1}^{K} [R(v^k)(v^k - \mu)^2]}{\sum_{k=1}^{K} R(v^k)}}, \tag{2.3}$$

where $v^k$ is $k^{th}$ sampled policy parameters and $R(v^k)$ denotes the reward for $v^k$. In the experiments, the reward was defined as the average swimming speed in one second period during steady swimming. A learning experiment was stopped when the following convergence criteria were reached: changes of the average swimming speed were within 2 mm/s for three successive episodes.

Although we did not need to specify the learning rate, number of sampled rollouts in one episode ($M$) and number of selected best samples ($K$) were hyperparameters that needed to be empirically tuned for different µBots (table S2). With the above setting, the experimental learning in general converged within 20 episodes (figure S1).

## 3. Results

### 3.1. Experimental motor learning for forward and backward swimming

For each robot, motor learning experiments were conducted for maximizing forward and backward swimming speed, from which we obtained optimized gaits after convergence in learning (movie S1). Each learning experiment was repeated three times with varying initial values of the CPG parameters, which converged to nearly identical speed (figure S1) and swimming gaits (figure S2) (the only exception was found for µBot-4 without the caudal-fin where multiple local optima were obtained, figure S1*d* for speed and S2*d* for gaits). For forward swimming, the optimized swimming speed increased as the DoFs increased, while the swimming speed in Body Length per second (BL/s) decreased from 1.3 BL/s (μBot-2) to 1.0 BL/s (μBot-4), and to 0.8 BL/s (μBot-6) (figure S3). Removing the caudal fin decreased forward speed significantly from 1 BL/s (173mm/s) to 0.37 BL/s (52 mm/s), confirming its critical role in thrust generation. For backward swimming, μBot-2 was unable to swim backward, while μBot-4 and μBot-6 reached maximal backward speed at 0.16 BL/s (27 mm/s) and 0.25 BL/s (55 mm/s), respectively (figure S3 and movie S2).

For each learning task, although the three repeated experiments converged to nearly identical speed and gaits, the CPG parameters determining intersegmental phases and intensities exhibited large differences upon convergence (in both means and variations, figure S4 and S5). The only exception was for the CPG parameter $\tau$ which also converged faster than other parameters. In terms of the learned motor inputs, the optimized frequencies in the three experiments were almost identical, while the intersegmental phases and intensities had large variations among the experiments (figure S6). These results showed that gait generation and swimming speed were dominated by motor frequency and suggested that they likely have low sensitivity to the intersegmental phases and intensities, both of which were investigated further and reported below.

**3.2. Frequency response – swimming gaits and speed**

We performed frequency response experiments on all three μBots (with caudal-fin) in forward swimming and examined the emergent swimming gaits and speed from underlying motor-controlled FSI (movie S3). This was done by changing frequency (adjusting $\tau$) of the corresponding CPG (1-11 Hz) while keeping other CPG parameters identical to those optimized. The swimming performance of μBots degraded quickly beyond the selected frequency range.

The caudal-fin tip velocity showed strong correlation to motor frequency (figure 2*a*) and peaked nearly at optimized frequencies for highest swimming speed (μBot-2: 7.3 Hz, μBot-4: 8.2 Hz, μBot-6: 8.8 Hz, figure 2*b*). This suggested that there was likely a resonance effect for the caudal fin. Caudal-fin tip velocity and swimming speed also exhibited similar trends with an approximately linear relationship in the high frequency (or speed) region (5.5-11 Hz, grey shadow, figure 2*c*), and a nonlinear relationship in the low frequency region (1-5.5 Hz, green shadow).

It was further found that the above two regions were associated with two modes of swimming gaits emerged from the motor-controlled FSI (movie S4). At the high frequency region, the swimming gaits of all three µBots resembled standing waves (figure 2*d*), i.e., the lateral displacements along the body long axis alternated between large amplitude (oscillatory points) and small amplitude (nodal points). Note that the nodal points did not necessarily coincide with the body mechanical joints (figure 2*e*). The head and caudal-fin tips were end oscillatory points that consistently exhibited large displacements. Intuitively, a nodal point can be considered as a virtual fulcrum point, and together with its two adjacent oscillatory points, composes a virtual lever. Therefore, the entire robot behaved as a chain of oscillating levers; each lever rotated around a virtual fulcrum, and together, they drove and overcame the hydrodynamic load on the distal caudal fin.

Since these gaits emerged within the high-frequency region including the resonant frequencies which are known to produce standing waves [27], we named these gaits *Standing Wave "Resonant" Gaits (SW "Res" Gaits)*. Note that with SW "Res" Gaits, swimming speed was proportional to caudal-fin tip velocity (figure 2*c*), and such linear predictability may underscore the dominance of the caudal-fin and the negligible contribution from the µBot body in thrust generation (see electronic supplementary material for detailed discussion).

At the low frequency region, the µBots exhibited backward traveling waves along their bodies (figure 2*f*), which resembled those commonly observed in fishes [1]. Therefore, we named these gaits *Traveling Wave "Fish-Like" Gaits* (*TW "FL" Gaits*). With TW "FL" Gaits, the swimming speed was no longer proportional to the caudal-fin tip velocity (figure 2*c*). However, as will be shown below, the linear predictability which existed for SW "Res" Gaits can be re-established by including the body undulatory velocity, which suggested the noneligible contribution to thrust generation from the µBot body that exhibited backward traveling waves, similar to those reported in fishes [28,29].

### 3.3. Invariant BCF Strouhal number ($St_{BCF}$)

We found that the averaged lateral undulatory velocity of all body segments and caudal fin combined (referred to as BCF lateral undulatory velocity, see electronic supplementary material), instead of caudal-fin tip velocity alone, proportionally predicted the swimming speed for all µBots operating at all tested frequencies (figure 3*a*). This led to an invariant Strouhal number ($St_{BCF}$ = 0.182, linear regression with coefficient of determination $R^2$ = 0.941), defined based on the undulation of body and caudal fin combined (see electronic supplementary material for details). Note that $R^2$ = 0.875 if the caudal-fin tip velocity was used for the linear regression.

We then calculated the $St_{BCF}$ for a collection of biological and bio-inspired robotic swimmers who's average BCF lateral undulatory velocity and swimming speed are available in the literature (figure 3*b* and table S3, see electronic supplementary material). Distribution of the $St_{BCF}$ is presented in figure 3*c*, where two clusters were clearly seen. Using regression clustering method [30] (see electronic supplementary material), we identified two regression lines in figure 3*b*, corresponding to two classes of undulatory swimming (with $St_{BCF}$ = 0.186 and $St_{BCF}$ = 0.066) among all biological and robotic swimmers examined here (with clustering loss $L_c$ = 0.08, electronic supplementary material). The line with $St_{BCF}$ = 0.186 closely matched the linear regression of µBots' data which gave $St_{BCF}$ = 0.182. Then we measured the silhouette score for the clustering (see electronic supplementary material). The average score (*AS*) of all data points was *AS* = 0.669. Note that the clustering became weakened if Strouhal number with merely caudal-fin tip velocity was used (figure S7, which gives $L_c$ = 0.249 and *AS* = 0.612). We referred to the class of $St_{BCF}$ = 0.186 as *slow* speed-to-undulation (swimming speed to BCF lateral undulation) swimmers and the class of $St_{BCF}$ = 0.066 as *fast* speed-to-undulation swimmers. In general, the examined biological thunniform and carangiform swimmers [10,31–34] belonged to the *fast* speed-to-undulation swimmers (dark blue, figure 3*b*), and the examined anguilliform biological swimmers [35–41] belonged to the *slow* speed-to-undulation swimmers (light blue) except needlefish [42], a fast anguilliform swimmer with lunate caudal-fin (similar to those of typical carangiform or thunniform swimmers), capable of swimming at 2 BL/s. Surprisingly, all the examined bio-inspired robotic swimmers, including µBots, belonged to the class of *slow* speed-to-undulation swimmers (grey), regardless of their morphological designs (e.g., anguilliform [43], carangiform [44–46] or thunniform [47,48]).

**3.4. Robustness of gaits to motor control disruptions**

Next, we tested whether the motor-controlled FSI of µBots possessed embodied robustness in generating TW "FL" Gaits and SW "Res" Gaits. This was done by applying disruptions (figure 4*a*) to the voltage signals (figure 4*b*), measuring the corresponding changes in the swimming speed and gaits (figure 4*c*), and quantifying their sensitivities (figure 4*d*) to the disruptions (see electronic supplementary material for details). Examples of disruption experiments and results are illustrated in movie S5.

The results for SW "Res" Gaits are shown in figure 4*e* and 4*f*. For all three µBots, the sensitivities of swimming speed to motor frequency were the highest (brown, figure 4*e*), while those to other motor control parameters were generally low. The only exception was for the intensity of the $2^{nd}$ actuator of µBot-2, which had mild effects on swimming speed (*in2*, green, Figure 4*ei*). The sensitivities of all gait parameters to motor frequency were also the highest (purple boxes, figure 4*f*), while those (except the head) to other motor control parameters were generally low (cyan boxes). Note that the head motion (including *m1* and *ip1* in figure 4*c*) was sensitive to both frequency and other motor control parameters (figure 4*f*), although it had no significant effects on the swimming speed.

The results for TW "FL" Gaits are shown in figure 4*g* and figure 4*h*. Compared to the results for SW "Res" Gaits, the swimming speed now became more sensitive to the motor intensities and intersegmental phases, although the frequency still played a dominating role. The swimming gaits remained sensitive to frequency in most cases, but with several exceptions (e.g., gait sensitivity to increasing frequency was low for µBot-6, *f+*, figure 4*hiii*). The gaits also became more sensitive to the motor intersegmental phases and intensities in general (e.g., more darker red squares in the cyan boxes were observed in figure 4*h*).

### 3.5. Frequency response – muscle power characteristics

We further examined the robot power characteristics in the frequency response by calculating actuators' work loops and cost of transport. The work loop is commonly used to evaluate the muscle physiology in terms of its mechanical work and power and can be obtained by plotting the time course of the muscle force and length, which forms a closed loop for rhythmic muscle contraction (the total enclosed area represents the net work output in one cycle) [49]. Here, we applied this method to µBots actuators (see electronic supplementary material). The work loops of µBots' actuators showed strong frequency dependency (figure 5*a*), as their shapes evolved substantially over the frequencies (see figure 5*b* for different functions of an actuator according to work loop shapes). At low frequencies (highlighted in green), all the actuators behaved as motors (large enclosed area for power output) combined with virtual masses (rightward inclined work loop for overcoming extrinsic elastic effects). At high frequencies (highlighted in yellow), the posterior actuators of µBot-4 (last 2) and µBot-6 (last 4) behaved as virtual springs (leftward inclined work loop) with low power output (small enclosed area); together with the rubber suit elasticity, they effectively stiffened the posterior body of the robots. For all three µBots, while the head actuators did not have a consistent function, the actuators behind the head exhibited the highest power output at high frequencies. Note that the diverse functions of the homogenous actuators of µBots are similar to those observed in biological muscles [50].

The total power output of all actuators combined in a µBot exhibited an inverted U-shaped dependency on the frequency (figure 5*c*). The optimized swimming speed occurred at frequencies higher than those for the maximum power output. We calculated two types of cost of transport: 1) net cost of transport, $CoT_{net}$, defined as the ratio of power output of a µBot's actuators to the swimming speed, and 2) weight-averaged net cost of transport, $CoT_{w-net}$, defined as $CoT_{net}$ divided by the weight of the µBot (which was analogous to the net metabolic CoT of fishes). The $CoT_{net}$ and $CoT_{w-net}$ curves also demonstrated inverted U-shapes (figure 5*d* and 5*e*). $CoT_{net}$ at the maximum speed for all µBots were similar (around 0.09 J m$^{-1}$). However, the $CoT_{w-net}$ at the maximum speed decreased substantially as the DoFs increased, with µBot-6 demonstrating the lowest $CoT_{w-net}$ (1.3 J m$^{-1}$ kg$^{-1}$). The $CoT_{w-net}$ of µBots is comparable with those reported for fishes with similar size (2.5 J m$^{-1}$ kg$^{-1}$ for bluegill sunfish [51]; 1.9 J m$^{-1}$ kg$^{-1}$ for mackerel and kawakawa [52]).

The above results suggested that the capability to transport mass was enhanced by adding body segments and DoFs. Although the current study is only limited to µBot with 6 body segments, it will be interesting to see if the $CoT_{w-net}$ at the maximum speed can be further reduced by adding more body segments, analogous to the improved fuel efficiency in a long rail freight trains with large number of freight cars.

## 4. Discussion

Our results based on the robophysical models (µBots) confirmed a number of embodied traits in motor-controlled FSI. First, the embodiment promotes fast convergence of motor control frequency and the emergence of diverse swimming gaits in µBots via simple frequency modulation. While the motor control signals are merely traveling waves, frequency modulation gives rise to a spectrum of swimming gaits ranging from traveling waves to standing waves. Second, despite the high sensitivity to motor frequency, the emergent swimming gaits remain robust against internal disruptions applied to motor signals such as intersegmental phases and intensities. Therefore, motor-controlled FSI adds another layer of robustness to gait generation in addition to those in the rhythmogenesis provided by a combination of feedforward control and peripheral sensory feedback [6]. The same study also underscores the importance of rhythm and rhythmogenesis in gait generation, consistent with the results of the current study. However, the motor control disruptions of our work were not as radical as those in their study. For example, our results showed that the intersegmental phases cannot be completely reversed (otherwise the robot will swim backward). Third, swimming speed of all µBots with varied DoFs and at all motor frequencies can be predicted simply by the respective BCF lateral undulatory velocity, which results in a single, invariant BCF Strouhal number in µBot swimming. Since Strouhal number is a dimensionless fluid number that characterizes the wake patterns and hydrodynamic traits of undulatory swimmers [12,53], our results show that the FSI in μBots' swimming possesses such invariant hydrodynamic properties that could be independent of morphological design, motor control and swimming gaits. Fourth, although the embodied traits in motor-controlled FSI point to simplicity in motor control, the homogenous actuators of μBots and the associated work loops demonstrate surprisingly diverse functions in terms of power output, virtual stiffness and mass, as they vary substantially along the body and under frequency modulation.

**4.1. Embodiment yields simplicity for control and high efferent predictability**

Collectively, the embodied traits identified in μBots suggest simplicity in both gait and speed control, as well as high predictability of motor efference (i.e., predicting movement [54], e.g., predicting swimming speed according to motor signals). Interestingly, the simplicity and predictability reside in a motor-controlled FSI system that has highly complex dynamics, which is difficult to be fully understood and accurately modelled [55]. Although a complex system is usually associated with unpredictability [56], our results suggest that there potentially exist simple properties embodied within the complex FSI underlying fish-inspired swimming. More importantly, these properties are easily "discoverable" by an external agent (e.g., via reinforcement learning), from which a simple predictive model from motor control to swimming performance can be established [57], although substantive work will be needed to extend the work to more robots design and animals data.

**4.2. FSI resonance in μBots likely enhances the embodied properties**

The aforementioned robustness and predictability in fish-inspired swimming are also likely enhanced by resonant effects in the FSI [58]. FSI resonance is clearly visible in the μBot caudal-fin from frequency response experiments – there are clear peaks of caudal-fin tip velocity at optimized frequencies (figure 2*a* and 2*b*). In addition, despite the motor control signals being traveling waves, the resultant SW "Res" Gaits are standing waves, which are commonly produced by resonant effects [27]. Notably, the SW "Res" Gaits are rarely observed in biological fishes. However, similar gaits of the standing-wave type have been reported previously in a simulation work based on models of fish swimming [59] and experimental studies with soft robots [60,61].

In sensitivity analysis (figure 4), it is evident that SW "Res" Gaits show a higher degree of robustness and stronger frequency modulation than those of TW "FL" Gaits. Since it is apparent that the former has stronger resonant effects than the latter, one can conclude that resonant effect enhances the robustness and frequency-modulation in the µBot gait generation.

The work loops of µBots (figure 5*a*) indicate that the resonance effects likely occur at their posterior body parts which show increasingly higher amount of virtual elastic storage with SW "Res" Gaits at the optimized frequencies. Specifically, with SW "Res" Gaits, the actuators at the anterior parts behave as motors, while those at the posterior parts behave approximately as virtual torsional springs. Note that an actuator, when behaving as a virtual spring in terms of its torque-displacement functioning, cannot physically store elastic energy as in physical springs, which instead will be dissipated as heat. Also note that for µBot-2, the posterior actuator does not show the spring effect since the elastic caudal-fin likely functions as the main source of elastic energy storage for resonance.

In addition, the FSI resonance is likely determined primarily by the effective body mass (including both robot physical mass and fluid added mass) and the effective body stiffness (including rubber suit physical stiffness and actuator virtual stiffness). For example, in authors' previous simulation work, it was shown that the fluid added-mass torque and effective spring torque (due to both physical and virtual stiffness) can balance each other in µBots' swimming [62]. In a separate study [63], it was shown, using analytical modelling and simulation, that the resonance can still occur even when the entire physical body mass and stiffness were removed, and the resonant frequency was likely determined by actuator virtual stiffness and fluid added mass. Nevertheless, the specific mechanisms that enable FSI resonance and the corresponding hydrodynamic phenomena may demand more detailed examinations using flow visualization and hydrodynamic analysis, which are beyond the scope of this work.

### 4.3. BCF Strouhal number and two fundamental classes of undulatory swimming

In this study, we show that the forward swimming of µBots can be better characterized by a constant BCF Strouhal number that considers both body and caudal-fin undulation, than the one with caudal-fin undulation alone. The BCF Strouhal number also enables us to categorize a large number of biological and robotic swimmers into only two classes of undulatory swimming (figure 3). These results suggest that body undulation may also contribute to propulsion, in addition to the caudal fin. Note that Lighthill concluded, based on mathematical modelling, that thrust is only related to caudal-fin movement, while body undulation affects propulsion efficiency [64,65]. In addition, we also demonstrated that body undulation alone can generate thrust, for example, in the backwards swimming and swimming without the caudal fin.

The two classes of undulatory swimming revealed by the BCF Strouhal number are *slow* speed-to-undulation and *fast* speed-to-undulation (figure 3). One common morphological trait that separates the two classes in biological fishes is the shape of the caudal-fins; the *fast* speed-to-undulation fishes have lunate shape caudal-fins, while *slow* speed-to-undulation fishes have pointed or rounded caudal-fins. However, all robotic swimmers examined in this study, regardless of the caudal-fin shape, fall into the *slow* speed-to-undulation class. This discrepancy suggests that there exist other design factors that influence the BCF undulatory propulsion in biological fishes that the existing robotic designs are unable to replicate – control strategy, dorsal fins, stiffness, etc.

So, what are these other factors contributing to the above discrepancy between robotic and biological swimmers? Although our work does not provide direct evidence to address this question, we can reject the following possibilities. First, our work shows that alternating the motor control does not affect the µBots' belonging to the *slow* speed-to-undulation class. The same is valid in biological fishes – within a species, varying swimming kinematics or speed does not affect the class that the species belongs to (it will only move the species along the same regression line in figure 3*b*). Second, the number of body segments (or the body slenderness) of µBots does not affect its class as well (figure 3*a*). Third, since fishes in both classes and some robots have dorsal fins, their presence is also unlikely to affect the classes, although dorsal fins have been shown to enhance the thrust generation in swimming [66]. Therefore, the most likely factor for robots to achieve *fast* speed-to-undulation class, as we speculate, could be the tuning of the spatial stiffness distribution in body and caudal-fin. It is known that fishes are able to actively tune their body stiffness to achieve resonance at various frequencies [67], and swimming performance is highly correlated to the body stiffness [68]. Compared with our previous work [20], the swimming performance of μBots can be improved substantially via body stiffness tuning. Nevertheless, future studies are required to fully address this question.

**Data accessibility**

All data generated are available in the paper or its electronic supplementary material.

**Author Contributions**

H.D. and B.C. conceived the study; H.D. prepared and conducted the experiments; H.D., D.L., and B.C. developed the data analysis; H.D. and B.C. wrote the first draft; All authors contributed to interpretation of the findings and revision of the manuscript.

**Acknowledgments**

We thank Dr. Shih-Jung Hsu for the preliminary contributions of the conceptual design of μBot and Patrick Burke for the help with μBot hardware. Funding for this research includes National Science Foundation CNS-1932130 (BC), Army Research Office W911NF-20-1-0226 (BC), and United States Department of Agriculture NIFA-2019-67021-28991 (SP).

**Competing Interests**

We declare we have no competing interests.

**References**

1. Sfakiotakis M, Lane DM, Davies JBC. 1999 Review of fish swimming modes for aquatic locomotion. *IEEE J. Ocean. Eng.* **24**, 237–252. (doi:10.1109/48.757275)

2. Chen JY, Huang DY, Li CW. 1999 An early Cambrian craniate-like chordate. *Nature* **402**, 518–522. (doi:10.1038/990080)

3. Videler JJ. 1993 *Fish Swimming*. London, U.K: Chapman and Hall.

4. Tytell ED, Hsu CY, Williams TL, Cohen AH, Fauci LJ. 2010 Interactions between internal forces, body stiffness, and fluid environment in a neuromechanical model of lamprey swimming. *Proc. Natl. Acad. Sci. U. S. A.* **107**, 19832–19837. (doi:10.1073/pnas.1011564107)

5. Chiel HJ, Beer RD. 1997 The brain has a body: Adaptive behavior emerges from interactions of nervous system, body and environment. *Trends Neurosci.* **20**, 553–557. (doi:10.1016/S0166-2236(97)01149-1)

6. Thandiackal R *et al.* 2021 Emergence of robust self-organized undulatory swimming based on local hydrodynamic force sensing. *Sci. Robot.* **6**, eabf6354.

7. Lauder G V, Drucker EG. 2004 Morphology and experimental hydrodynamics of fish fin control surfaces. *IEEE J. Ocean. Eng.* **29**, 556–571. (doi:10.1109/JOE.2004.833219)

8. Di Santo V, Goerig E, Wainwright DK, Akanyeti O, Liao JC, Castro-Santos T, Lauder G V. 2021 Convergence of undulatory swimming kinematics across a diversity of fishes. *Proc. Natl. Acad. Sci. U. S. A.* **118**, e2113206118. (doi:10.1073/pnas.2113206118/-/DCSupplemental.Published)

9. Bainbridge R. 1958 The speed of swimming of fish as related to size and to the frequency and amplitude of the tail beat. *J. Exp. Biol.* **35**, 109–133.

10. Dickson KA, Donley JM, Hansen MW, Peters JA. 2012 Maximum sustainable speed, energetics and swimming kinematics of a tropical carangid fish, the green jack Caranx caballus. *J. Fish Biol.* **80**, 2494–2516. (doi:10.1111/j.1095-8649.2012.03302.x)

11. Rohr JJ, Fish FE. 2004 Strouhal numbers and optimization of swimming by odontocete cetaceans. *J. Exp. Biol.* **207**, 1633–1642. (doi:10.1242/jeb.00948)

12. Taylor GK, Nudds RL, Thomas ALR. 2003 Flying and swimming animals cruise at a Strouhal number tuned for high power efficiency. *Nature* **425**, 707–711.

13. Saadat M, Fish FE, Domel AG, Di Santo V, Lauder G V., Haj-Hariri H. 2017 On the rules for aquatic locomotion. *Phys. Rev. Fluids* **2**, 1–12. (doi:10.1103/PhysRevFluids.2.083102)

14. Blight AR. 1977 The muscular control of vertebrate swimming movements. *Biol. Rev.* **52**, 181–218.

15. Cianchetti M, Follador M, Mazzolai B, Dario P, Laschi C. 2012 Design and development of a soft robotic octopus arm exploiting embodied intelligence. In *2012 IEEE International Conference on Robotics and Automation*, pp. 5271–5276. IEEE. (doi:10.1109/ICRA.2012.6224696)

16. Jiao Y, Ling F, Heydari S, Heess N, Merel J, Kanso E. 2021 Learning to swim in potential flow. *Phys. Rev. Fluids* **6**, 1–20. (doi:10.1103/PhysRevFluids.6.050505)

17. Snadwick RE, Sieffensen JF, Katz SL, Knower T. 1998 Muscle dynamics in fish during

steady swimming. *Am. Zool.* **38**, 755–770. (doi:10.1093/icb/38.4.755)

18. Aguilar J *et al.* 2016 A review on locomotion robophysics: The study of movement at the intersection of robotics, soft matter and dynamical systems. *Reports Prog. Phys.* **79**, 110001.

19. Gravish N, Lauder G V. 2018 Robotics-inspired biology. *J. Exp. Biol.* **221**. (doi:10.1242/jeb.138438)

20. Deng H, Burke P, Li D, Cheng B. 2021 Design and experimental learning of swimming gaits for a magnetic, modular, undulatory robot. In *2021 IEEE/RSJ International Conference on Intelligent Robots and Systems (IROS)*, pp. 9562–9568. IEEE. (doi:10.1109/IROS51168.2021.9636100)

21. Deng H, Li D, Panta K, Cheng B. 2023 Magnetic, modular, undulatory robots as robophysical models for exploration of fish-inspired swimming. *Bull. Am. Phys. Soc.*

22. Akanyeti O, Di Santo V, Goerig E, Wainwright DK, Liao JC, Castro-Santos T, Lauder G V. 2022 Fish-inspired segment models for undulatory steady swimming. *Bioinspiration and Biomimetics* **17**. (doi:10.1088/1748-3190/ac6bd6)

23. Ijspeert AJ. 2008 Central pattern generators for locomotion control in animals and robots: A review. *Neural Networks* **21**, 642–653.

24. Li L, Nagy M, Graving JM, Bak-Coleman J, Xie G, Couzin ID. 2020 Vortex phase matching as a strategy for schooling in robots and in fish. *Nat. Commun.* **11**, 1–9. (doi:10.1038/s41467-020-19086-0)

25. Matsuoka K. 1985 Sustained oscillations generated by mutually inhibiting neurons with adaptation. *Biol. Cybern.* **52**, 367–376. (doi:10.1007/BF00449593)

26. Wang J, Uchibe E, Doya K. 2016 EM-based policy hyper parameter exploration: application to standing and balancing of a two-wheeled smartphone robot. *Artif. Life Robot.* **21**, 125–131. (doi:10.1007/s10015-015-0260-7)

27. Halliday D, Resnick R, Walker J. 2013 *Fundamentals of Physics*. John Wiley & Sons, New York.

28. Gemmell BJ, Fogerson SM, Costello JH, Morgan JR, Dabiri JO, Colin SP. 2016 How the bending kinematics of swimming lampreys build negative pressure fields for suction thrust. *J. Exp. Biol.* **219**, 3884–3895. (doi:10.1242/jeb.144642)

29. Du Clos KT, Dabiri JO, Costello JH, Colin SP, Morgan JR, Fogerson SM, Gemmell BJ. 2019 Thrust generation during steady swimming and acceleration from rest in anguilliform swimmers. *J. Exp. Biol.* **222**, jeb212464.

30. Zhang B. 2003 Regression clustering. In *Proceedings of the Third IEEE International Conference on Data Mining*, pp. 451–458. (doi:10.1109/icdm.2003.1250952)

31. Jayne BC, Lauder G V. 1995 Speed effects on midline kinematics during steady undulatory swimming of largemouth bass, Micropterus salmoides. *J. Exp. Biol.* **198**, 585–602.

32. Dowis HJ, Sepulveda CA, Graham JB, Dickson KA. 2003 Swimming performance studies on the eastern Pacific bonito Sarda chiliensis, a close relative of the tunas (family Scombridae): II. Kinematics. *J. Exp. Biol.* **206**, 2749–2758. (doi:10.1242/jeb.00496)

33. Donley JM, Shadwick RE. 2003 Steady swimming muscle dynamics in the leopard shark Triakis semifasciata. *J. Exp. Biol.* **206**, 1117–1126. (doi:10.1242/jeb.00206)

34. Dewar H, Graham J. 1994 Studies of tropical tuna swimming performance in a large water tunnel III. kinematics. *J. Exp. Biol.* **192**, 45–59. (doi:10.1242/jeb.192.1.45)

35. D'Août K, Aerts P. 1999 A kinematic comparison of forward and backward swimming in the eel Anguilla anguilla. *J. Exp. Biol.* **202**, 1511–1521.

36. Herrel A, Choi HF, De Schepper N, Aerts P, Adriaens D. 2011 Kinematics of swimming in two burrowing anguilliform fishes. *Zoology* **114**, 78–84. (doi:10.1016/j.zool.2010.10.004)

37. Gillis GB. 1998 Environmental effects on undulatory locomotion in the American eel Anguilla rostrata: kinematics in water and on land. *J. Exp. Biol.* **201**, 949–961.

38. Lim JL, Winegard TM. 2015 Diverse anguilliform swimming kinematics in Pacific hagfish (Eptatretus stoutii) and Atlantic hagfish (Myxine glutinosa). *Can. J. Zool.* **93**, 213–223. (doi:10.1139/cjz-2014-0260)

39. Tack NB, Du Clos KT, Gemmell BJ. 2021 Anguilliform locomotion across a natural range of swimming speeds. *Fluids* **6**, 127. (doi:10.3390/FLUIDS6030127)

40. Gillis GB. 1997 Anguilliform locomotion in an elongate salamander (Siren intermedia): effects of speed on axial undulatory movements. *J. Exp. Biol.* **200**, 767–784. (doi:10.1242/jeb.200.4.767)

41. D'Août K, Aerts P. 1997 Kinematics and efficiency of steady swimming in adult axolotls (Ambystoma mexicanum). *J. Exp. Biol.* **200**, 1863–1871. (doi:10.1242/jeb.200.13.1863)

42. Liao JC. 2002 Swimming in needlefish (Belonidae): anguilliform locomotion with fins. *J. Exp. Biol.* **205**, 2875–2884. (doi:10.1242/jeb.205.18.2875)

43. Porez M, Boyer F, Ijspeert AJ. 2014 Improved lighthill fish swimming model for bio-inspired robots: modeling, computational aspects and experimental comparisons. *Int. J. Rob. Res.* **33**, 1322–1341. (doi:10.1177/0278364914525811)

44. Epps BP, Valdivia Y Alvarado P, Youcef-Toumi K, Techet AH. 2009 Swimming performance of a biomimetic compliant fish-like robot. *Exp. Fluids* **47**, 927–939. (doi:10.1007/s00348-009-0684-8)

45. Zhong Y, Song J, Yu H, Du R. 2018 A study on kinematic pattern of fish undulatory locomotion using a robot fish. *J. Mech. Robot.* **10**, 041013. (doi:10.1115/1.4040434)

46. Clapham RJ, Hu H. 2014 ISplash-I: High performance swimming motion of a carangiform robotic fish with full-body coordination. In *2014 IEEE International Conference on Robotics and Automation (ICRA)*, pp. 322–327. IEEE. (doi:10.1109/ICRA.2014.6906629)

47. Zhu J, White C, Wainwright DK, Di Santo V, Lauder G V., Bart-Smith H. 2019 Tuna robotics: A high-frequency experimental platform exploring the performance space of swimming fishes. *Sci. Robot.* **4**, eaax4615. (doi:10.1126/scirobotics.aax4615)

48. White CH, Lauder G V, Bart-Smith H. 2021 Tunabot Flex: a tuna-inspired robot with body flexibility improves high-performance swimming. *Bioinspir. Biomim.* **16**, 026019. (doi:10.1088/1748-3190/abb86d)

49. Josephson RK. 1985 Mechanical power output from striated muscle during cyclic

contraction. *J. Exp. Biol.* **114**, 493–512. (doi:10.1242/jeb.114.1.493)

50. Dickinson MH, Farley CT, Full RJ, Koehl MAR, Kram R, Lehman S. 2000 How animals move: An integrative view. *Science* **288**, 100–106. (doi:10.1126/science.288.5463.100)

51. Kendall JL, Lucey KS, Jones EA, Wang J, Ellerby DJ. 2007 Mechanical and energetic factors underlying gait transitions in bluegill sunfish (Lepomis macrochirus). *J. Exp. Biol.* **210**, 4265–4271. (doi:10.1242/jeb.009498)

52. Sepulveda C, Dickson KA. 2000 Maximum sustainable speeds and cost of swimming in juvenile kawakawa tuna (Euthynnus affinis) and chub mackerel (Scomber japonicus). *J. Exp. Biol.* **203**, 3089–3101. (doi:10.1242/jeb.203.20.3089)

53. Gazzola M, Argentina M, Mahadevan L. 2014 Scaling macroscopic aquatic locomotion. *Nat. Phys.* **10**, 758–761. (doi:10.1038/nphys3078)

54. Grillner S, El Manira A. 2020 Current principles of motor control, with special reference to vertebrate locomotion. *Physiol. Rev.* **100**, 271–320. (doi:10.1152/physrev.00015.2019)

55. Ijspeert AJ. 2014 Biorobotics: Using robots to emulate and investigate agile locomotion. *Science* **346**, 196–203.

56. Goldenfeld N, Kadanoff LP. 1999 Simple lessons from complexity. *Science* **284**, 87–89. (doi:10.1126/science.284.5411.87)

57. Bialek W, Nemenman I, Tishby N. 2001 Predictability, complexity, and learning. *Neural Comput.* **13**, 2409–2463.

58. Maurice P, Hogan N, Sternad D. 2018 Predictability, force, and (anti)resonance in complex object control. *J. Neurophysiol.* **120**, 765–780. (doi:10.1152/jn.00918.2017)

59. Gazzola M, Argentina M, Mahadevan L. 2015 Gait and speed selection in slender inertial swimmers. *Proc. Natl. Acad. Sci. U. S. A.* **112**, 3874–3879. (doi:10.1073/pnas.1419335112)

60. Wolf Z, Lauder G V. 2022 A fish-like soft-robotic model generates a diversity of swimming patterns. *Integr. Comp. Biol.* **00**, 1–14. (doi:10.1093/icb/icac039)

61. Jusufi A, Vogt DM, Wood RJ, Lauder G V. 2017 Undulatory swimming performance and body stiffness modulation in a soft robotic fish-inspired physical model. *Soft Robot.* **4**, 202–210. (doi:10.1089/soro.2016.0053)

62. Li D, Deng H, Bayiz YE, Cheng B. 2022 Effects of Design and Hydrodynamic Parameters on Optimized Swimming for Simulated, Fish-inspired Robots. In *2022 IEEE International Conference on Intelligent Robots and Systems (IROS)*, pp. 7500–7506. IEEE. (doi:10.1109/IROS47612.2022.9981478)

63. Kohannim S, Iwasaki T. 2014 Analytical insights into optimality and resonance in fish swimming. *J. R. Soc. Interface* **11**, 20131073. (doi:10.1098/rsif.2013.1073)

64. Lighthill MJ. 1960 Note on the swimming of slender fish. *J. Fluid Mech.* **9**, 305–317.

65. Lighthill MJ. 1971 Large-amplitude elongated-body theory of fish locomotion. In *the Royal Society of London. Series B: Biological Sciences*, pp. 125–138.

66. Drucker EG, Lauder G V. 2005 Locomotor function of the dorsal fin in rainbow trout:

Kinematic patterns and hydrodynamic forces. *J. Exp. Biol.* **208**, 4479–4494. (doi:10.1242/jeb.01922)

67. McHenry MJ, Pell CA, Long Jr JH. 1995 Mechanical control of swimming speed: stiffness and axial wave form in undulating fish models. *J. Exp. Biol.* **198**, 2293–2305.

68. Wang T, Ren Z, Hu W, Li M, Sitti M. 2021 Effect of body stiffness distribution on larval fish – like efficient undulatory swimming. *Sci. Adv.* **7**, eabf7364.

**Figures**

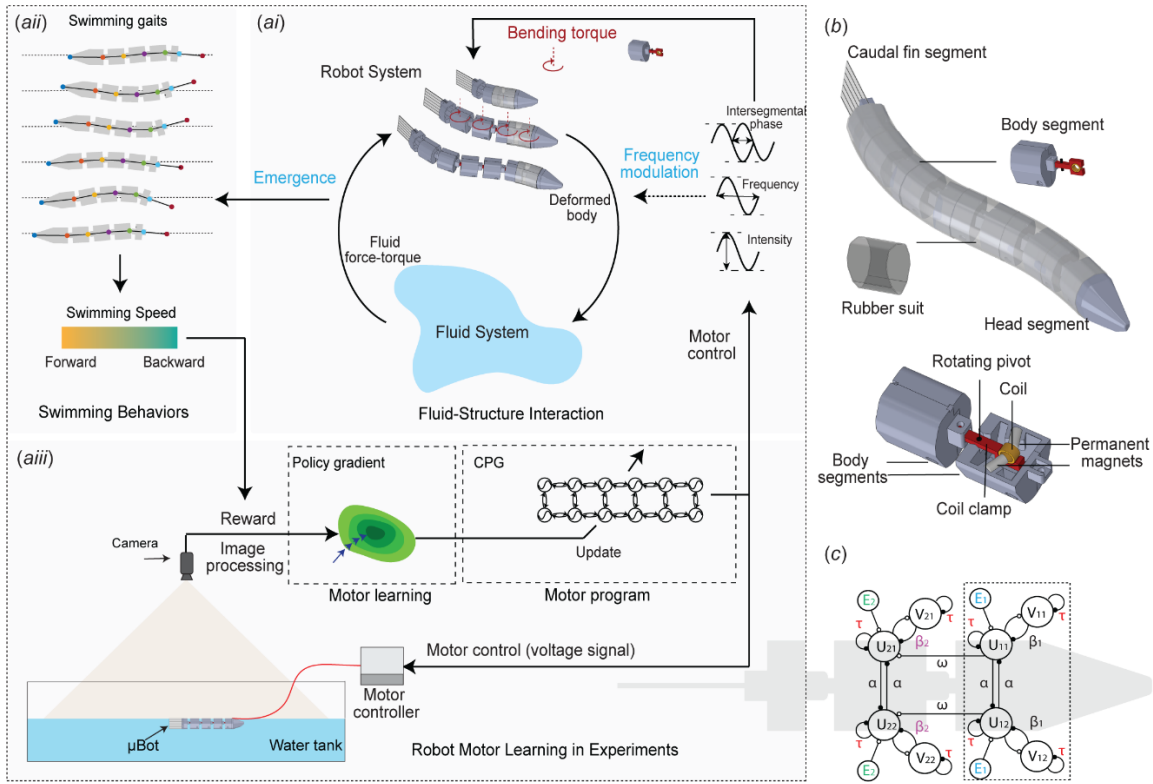

**Figure 1.** Overview of µBot motor learning experiment and µBot design. (*a*) Framework of µBot motor learning experiments for swimming, which includes (i) the Fluid-Structure Interaction (FSI) between the robot and fluid system; (ii) the emergence of swimming gaits and performance; and (iii) the experimental setup for robot learning, where the FSI is motor-controlled according to a motor program (CPG) which was optimized via policy-gradient RL method. (*b*) Design of µBots (µBot-6 used as an example) with a cross-section view of the body segment. (*c*) µBot motor program designed as a CPG with its parameters labelled (µBot-2 used as an example). Dashed box represents a CPG module corresponding to a single actuator, which includes two neurons inhibiting each other. Parameters labelled with colours were learned experimentally, while those in black were fixed *a priori*.

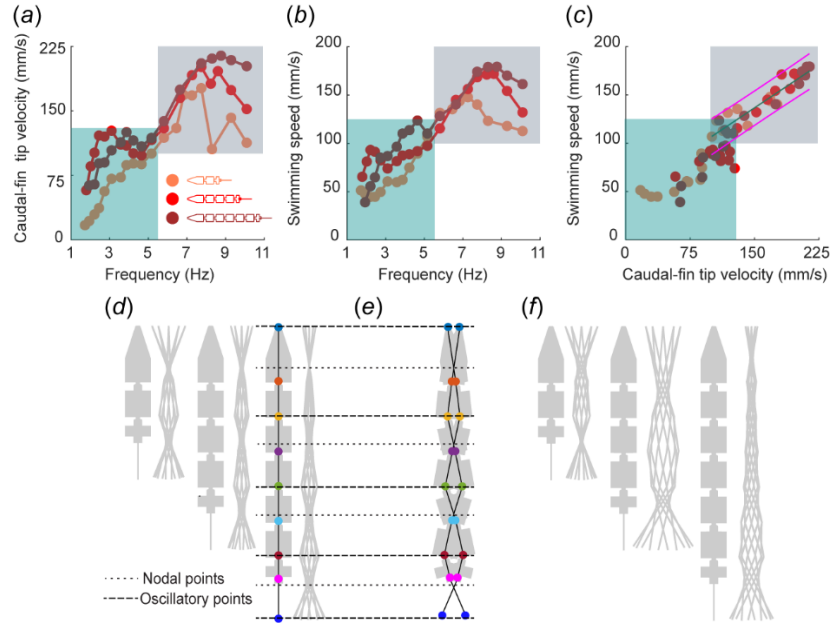

**Figure 2.** (*a*) Dependency of caudal-fin tip velocity on motor frequency. (*b*) Dependency of swimming speed on motor frequency. (*c*) Relationship between swimming speed and caudal-fin tip velocity. At the high frequency region (grey shadow, 5.5-11 Hz), there was a linear relationship between the swimming speed and the caudal-fin tip velocity, as illustrated by the linear regression (green line) and the 95% prediction intervals (pink lines). However, at the low frequency region (green shadow, 1-5.5 Hz), the linear relationship did not hold. The two regions also corresponded to the two modes of swimming gaits presented in (*d*) and (*f*). (*d*) Midline kinematics of high frequency Standing-Wave "Resonant" Gaits (SW "Res" Gaits) at optimized frequency (7.3, 8.2, 8.8 Hz for µBot-2, µBot-4, and µBot-6, respectively). (*e*) Illustration of SW "Res" Gaits. The robot joints are marked as filled colour dots, and the gait nodal points and oscillatory points are illustrated as dotted lines and dashed lines, respectively. The oscillatory points consistently coincided with the robot joints, which was not the case for nodal points. (*f*) Midline kinematics of Traveling-Wave "Fish-Like" Gaits (TW "FL" Gaits) at low frequency (3.1, 2.1, 2.7 Hz for µBot-2, µBot-4, and µBot-6, respectively). The lateral displacements of all the midline kinematics are scaled up two times for clearer illustrations. Kinematic snapshots in (*d*) and (*f*) show the displacement of the body midline at 10 equally spaced time intervals during a single tail beat cycle.

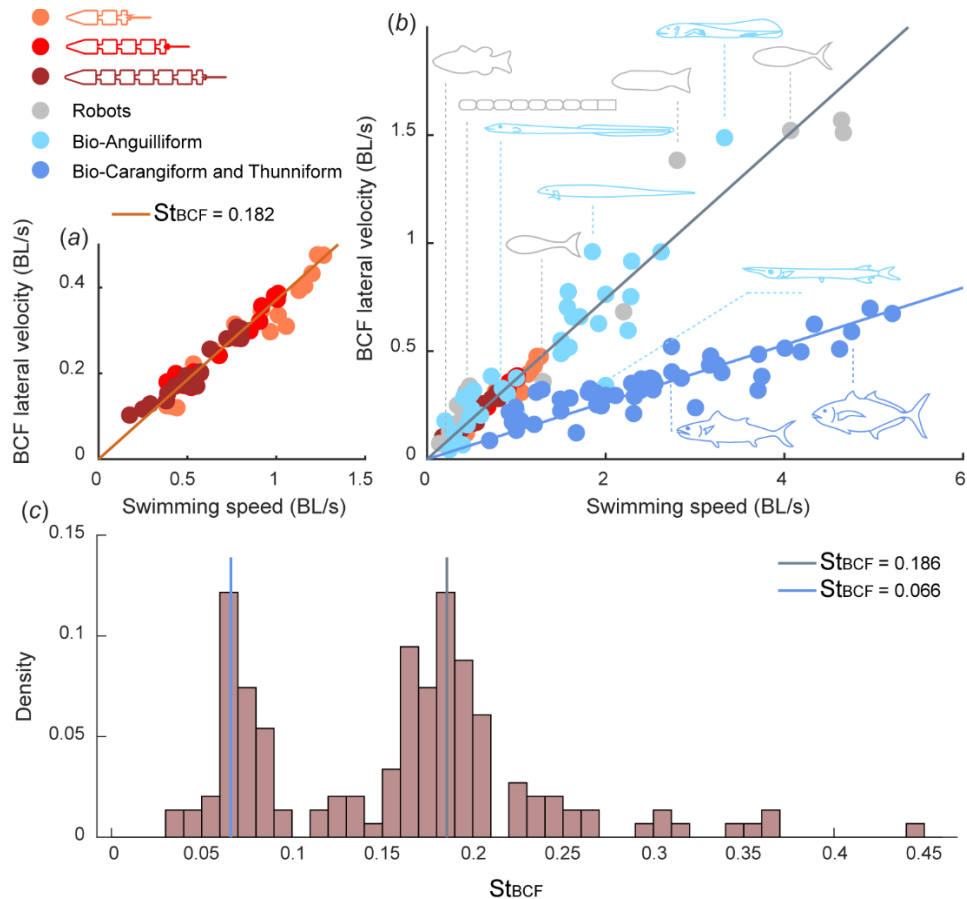

**Figure 3.** (*a*) Relationship between swimming speed and average BCF lateral undulatory velocity for μBots with linear regression (solid line). (*b*) Relationship between swimming speed and average BCF lateral undulatory velocity for μBots, other robotic fishes, and biological fishes surveyed from the literature. The solid lines were acquired from the regression clustering method, the slope of which corresponds to twice the values of $St_{BCF}$. The sketches illustrate the body shapes of some fishes and robots as examples (not to scale). (*c*) Histogram of the distribution of $St_{BCF}$ with each column representing the density of the data falling within the range. The two vertical lines mark the values of $St_{BCF}$ from the regression clustering.

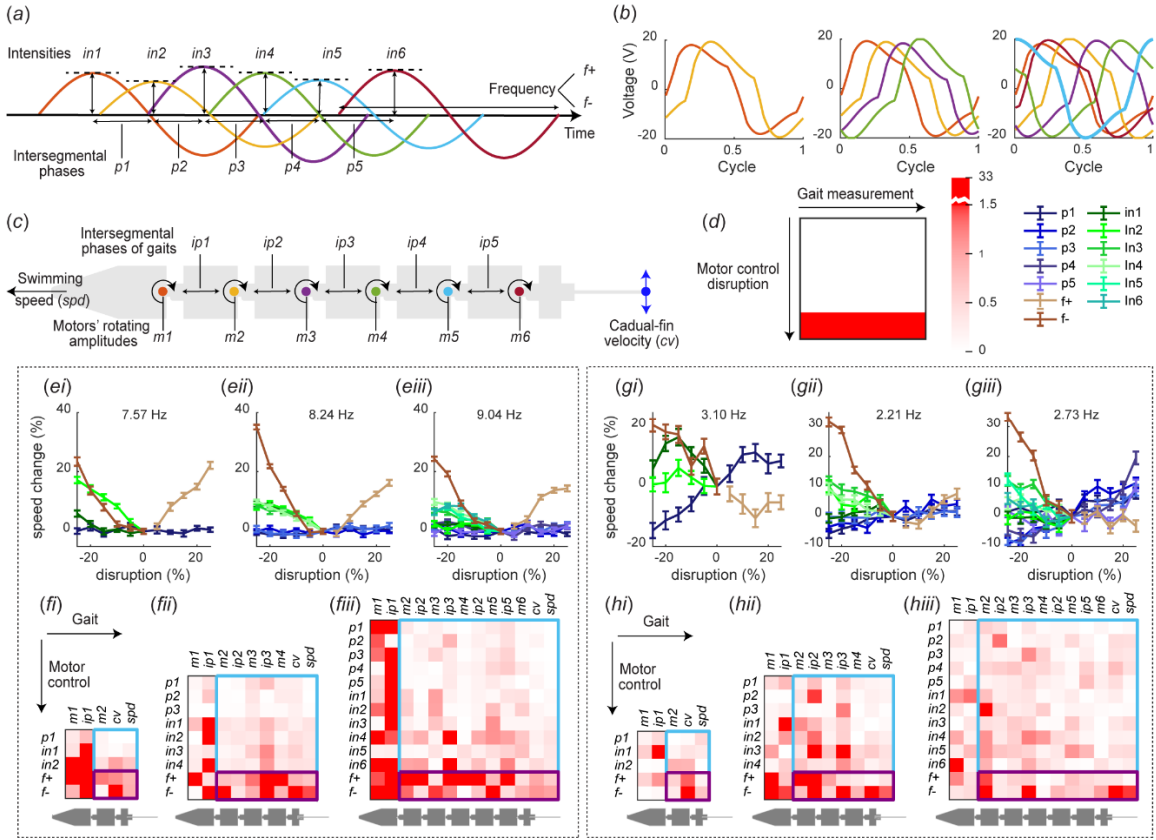

**Figure 4.** (*a*) Characteristics of the motor signals for μBot-6. μBot-2 and μBot-4 use only the first 2 and 4 signals, respectively. *in1* to *in6* represent the intensities of the motor signals. *p1* to *p5* represent the intersegmental phases between motor signals applied to neighbouring actuators. *f+* and *f-* represent the increase and decrease of the frequency, respectively. (*b*) Nominal motor signal inputs for μBot-2, μBot-4, μBot-6. (*c*) Gait parameters to be measured. *ip1* to *ip6* represent the intersegmental phases between the neighbouring joint angles. *m1* to *m6* represent the amplitude of joint angles. *spd* and *cv* are abbreviations for swimming speed and caudal-fin tip velocity, respectively. (*d*) Examples for presenting the result of sensitivity analysis and legends for the following plots. The sensitivity magnitudes are indicated by the colour bar. Note that the magnitudes above 1.5 are all considered maximum sensitivity. (*e* and *f*) Sensitivity of swimming speed (*e*) and gaits (*f*) to various disruptions applied to the motor control of SW "Res" Gaits for μBot-2, μBot-4, μBot-6, respectively. (*g* and *h*) Sensitivity of swimming speed (*g*) and gaits (*h*) to various disruptions

applied to the motor control for TW "FL" Gaits for µBot-2, µBot-4, µBot-6, respectively. Error bars indicate the standard deviation in (*e*) and (*g*).

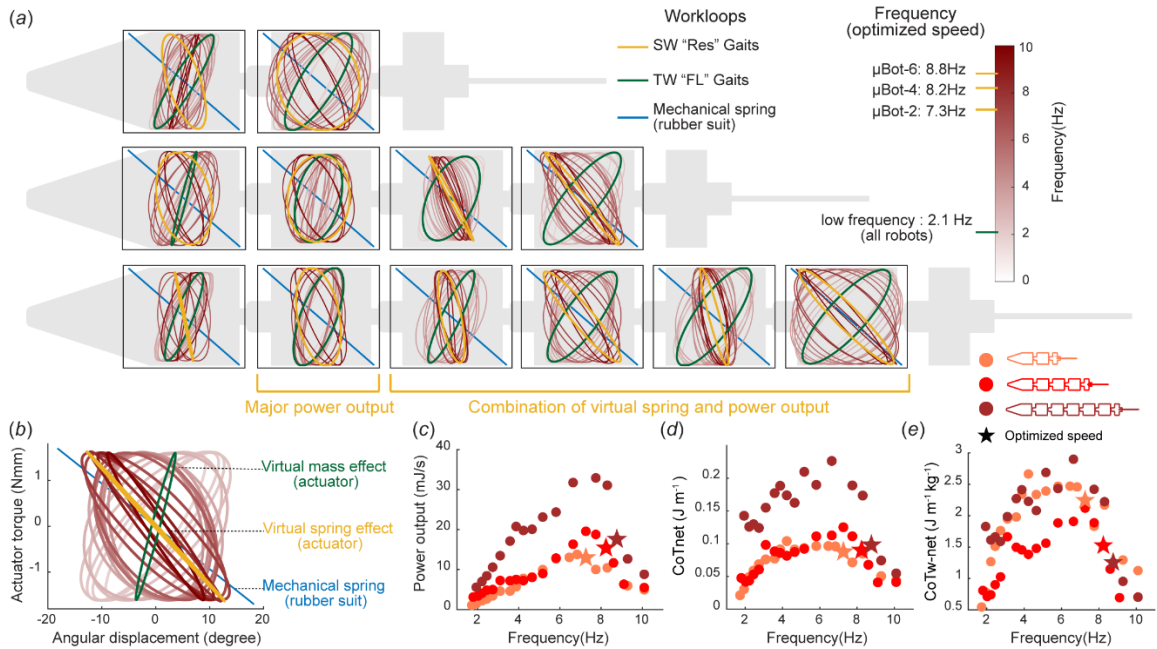

**Figure 5.** (*a*) Changes in work loop shapes according to motor control frequency. The work loops of SW "Res" Gaits and TW "FL" Gaits are highlighted in yellow and green, respectively. (*b*) Illustration of work loops corresponding to mechanical springs, virtual springs and virtual masses. (*c*), (*d*), and (*e*) Dependencies of total actuator power output, net cost of transport CoT$_{net}$, and weight-averaged net cost of transport CoT$_{w\text{-}net}$ on motor frequency, respectively. The frequencies for optimized speed are marked with stars.